%% file: samplepaper.tex
\documentclass[runningheads]{llncs}
\usepackage{graphicx}
\usepackage{booktabs}
\usepackage{subfigure}
\usepackage[utf8x]{inputenc}
\usepackage{appendix}
\usepackage{multirow}
\usepackage{color}

\newcommand\clement[1]{\textcolor{red}{#1}}
\newcommand\julien[1]{\textcolor{blue}{#1}}
\definecolor{jc}{rgb}{0,.66,.4}

\begin{document}
\title{How to detect novelty in textual data streams? A comparative study of existing methods}
\titlerunning{A comparative study of novelty detection methods}
%

\author{Clément Christophe\inst{1,2} \and
Julien Velcin\inst{1} \and
Jairo Cugliari\inst{1} \and
Philippe Suignard \inst{2} \and
Manel Boumghar \inst{2}
}
\authorrunning{C. Christophe et al.}
%
\institute{Université de Lyon, Lyon 2, ERIC, France \email{{\{julien.velcin, jairo.cugliari\}@univ-lyon2.fr}} \and EDF R\&D, Palaiseau, France \email{\{clement.christophe,philippe.suignard,manel.boumghar\}@edf.fr}}
\maketitle              
\begin{abstract}
Since datasets with annotation for novelty at the document and/or word level are not easily available, we present a simulation framework that allows us to create different textual datasets in which we control the way novelty occurs. We also present a benchmark of existing methods for novelty detection in textual data streams. We define a few tasks to solve and compare several state-of-the-art methods. The simulation framework allows us to evaluate their performances according to a set of limited scenarios and test their sensitivity to some parameters. Finally, we experiment with the same methods on different kinds of novelty in the New York Times Annotated Dataset.

\keywords{Novelty Detection \and Text mining \and Evaluation framework \and Natural Language Processing}
\end{abstract}

\section{Introduction}

This work has been mainly driven by an industrial use case, in which our aim is to detect as soon as possible emerging issues in data related to customer returns (e.g., emails, complaints). ``Ranking plans'' (meaning, predefined categories) are already used to classify these data but the partners have realized that classes can evolve over time. This raises the problem of detecting new classes (our ``novelty'') in textual data streams.

It is a complex issue that can be related to several domains (e.g., signal processing, data mining), and which, because of its usefulness and its many fields of application, is particularly interesting to study. Although this issue has attracted attention in several critical areas: video-surveillance, physiological monitoring, cancer detection, immunology, its development in textual data is still limited outside of event detection \cite{ng2007novelty,yang2002topic}.
The lack of development of these approaches at the level of textual data can be due to the specific nature of these data type and the difficulty of choosing the best representation for the task. It is common to use topic modeling techniques to represent textual data, such as LDA \cite {blei2003latent} or its temporal variants \cite{blei2006dynamic,gerrish2010language}. These methods are particularly used to study the appearance of concept drifts \cite{suzuki2014detection,murena2018adaptive}.

Different applications consider novelty in different ways: some compare it to a single time event and therefore use event detection methods \cite{metzler2012structured}, others define it as an outlier detection task \cite{ritter1997outliers} or first story detection (FSD) task \cite{allan2000detections}.
Besides, these methods are rarely tested on the same types of data (e.g., Twitter data, scientific abstract, press articles).
In this work, we aim at comparing different families of method on a common dataset, and on different tasks and controlled scenarios.

It turns out that, after studying the literature on novelty detection, we notice that despite the fact that the general idea remains more or less the same, the formalism is not identical depending on the application cases.
We notice few studies that analyze different methods of novelty detection. The work of Markou and Singh \cite{markou2003novelty1} distinguishes between statistical approaches and neural network approaches, which may not be very modern given the convergence of these two domains.
Marsland's work \cite{marsland2003novelty} gives us a description of the novelty in biological organisms and points out the lack of precise definition in the literature. Finally, the work of Pimentel \cite{pimentel2014review} allows the creation of a taxonomy related to methods of novelty detection; it classifies the approaches according to 5 major categories. (1) The probabilistic approach tries to estimate the density of the ``normal'' class. (2) The distance-based approach includes nearest-neighbour and clustering techniques. (3) The reconstruction-based approach involves a regression model that would fit the training data with the intuition that a high reconstruction error is a sign of novelty. (4) The domain-based approach consists of defining a boundary between ``normal'' and ``abnormal'' data. Finally, (5) the information-theoretic techniques are based on the idea that novel data will significantly alter the information content in a dataset. Here, we want to apply novelty detection methods to textual data streams. We realized that even if some works exist today, we can say that there is no unique definition of novelty and that these methods regularly use information in addition to plain text (e.g., ReTweet relation in Twitter, citation network).

In order to shed light on the available approaches, we study novelty detection by defining a common framework. In other words, we use a general definition of novelty that allows us to define specific tasks to solve. Then, we select a number of previous works addressing tasks close to novelty detection and extend them to solve our tasks. We perform experiments to determine which methods are the most effective in which cases, and what are the parameters that influence the performance. The rest of our work is organized as follows. In Section \ref{simulation}, we describe our simulation framework. In Section \ref{experiments} we give information on the benchmark methodology and on the tested algorithms.
Finally, in Section \ref{results} we present our results on simulated datasets and on the New York Times Annotated Dataset.

\section{Definition of Novelty} \label{definition}

In this section, we review the definitions adopted in the literature before bringing them together into a common framework.

\subsection{Novelty in the literature}

The term ``novelty'' is often referred to ``anomalies'' or ``outliers'', but all these refer to a fixed observation over time. The novelty to detect is related to a signal: a novelty is an abnormal or unexpected change in a signal, which means that the observed state is different from what we can expect. It can be defined to deal with various types of signal: electronic systems, network logs, medical diagnostics, monitoring of industrial equipment or textual data. In each of these cases, the novelty is materialized by a change in the nature of the signal that is considered as abnormal or unexpected. It is therefore necessary to pay attention to the temporal aspect of this signal in order to characterize it. This characterization depends on what we want to detect concretely. In order to evaluate our methods, we have to structure the different novelty types.
For textual data, the word or topic frequency over time can stand for the signal we want to monitor.

\begin{figure}[h!]
\centering
\includegraphics[width=0.8\textwidth]{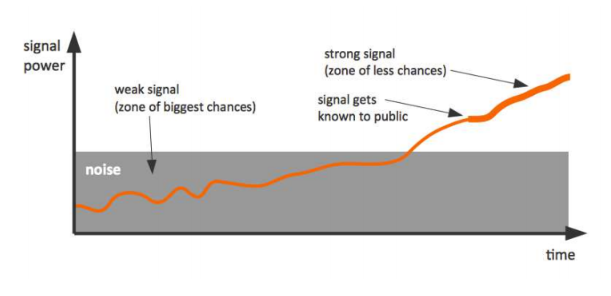}
\caption{\label{fig:formalisation}Evolution of a weak signal over time. \cite{eckhoff2014detecting}}
\end{figure}

One type of novelty we want to detect corresponds to weak signals whose frequency increases over time before becoming strong signals. It is a distinction between novelty and noise: the novelty becomes a strong signal while the noise remains a weak signal that has little interest for us. The work of \cite{hiltunen2010weak} gives us a definition of a weak signal: a weak ``new'' signal must come or be recognized by an expert group (see Figure~\ref{fig:formalisation}). This signal anticipates phenomena having impacts on the future and may include features that must be detected as soon as possible. A new weak signal is growing by itself over time and it is an early warning of an emerging trend. Mannermaa \cite{mannermaa2004heikoista} defines a weak signal as a phenomenon that is unlikely to appear but has a large potential for influence. These two definitions are not clear enough to develop an automatic detection system. Actually, it is mentioned that a weak signal needs to be recognized by a group of experts. This does not remove the human from the loop.\\

A novelty can also correspond to the sudden increase of a signal during a very short time: we observe a ``peak''.
Another type of novelty may correspond to an unusual change in frequency. Actually, the usual periodicity of a signal can be modified and this can be considered as a novelty. Finally, the analysis of concept drifts can also be considered as a novelty detection task. In these works \cite{suzuki2014detection,murena2018adaptive}, the methods aim at detecting a change in the parameters of the targeted classes.

\subsection{Our definition of novelty}

\begin{figure}[h!]
\centering
\includegraphics[width=\textwidth]{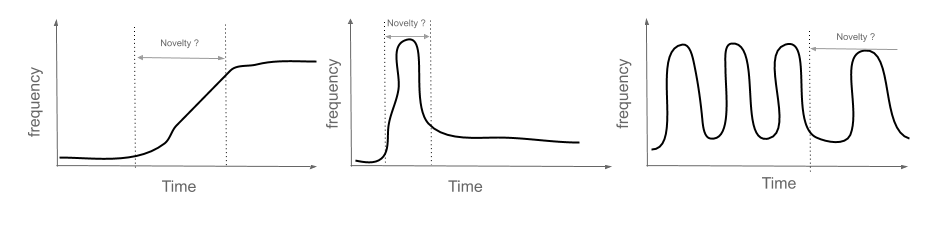}
\caption{\label{fig:types}Three possible types of novelty.}
\end{figure}

We now present what we consider as a novelty, i.e. what signal types can potentially interest us. Figure \ref{fig:types} summarizes the 3 types we focus on in this paper. We find the type of weak signal that evolves into a strong signal slowly over time (the most interesting one for our industrial use case), the event type signal that shows a sudden burst, and finally the cyclical signal.

The cyclical signal is not necessarily considered as a novelty. Actually, it depends on the users and what they want to detect.
However, we choose to keep it nonetheless for broadening the scope of our experiments. The main question remains the same: from when can we consider a signal as being novel? Finally, we mention the definition of \cite{mannermaa2004heikoista} that introduces the concept of influence. Several methods in the literature model this influence \cite{gerrish2010language} by applying it to data from scientific articles. We realized that this influence was not synonymous with novelty. Actually, some scientific papers can be extremely quoted and influential for a field without being the first innovative method of a domain. This explains why we choose not to consider the structural information provided by the links between documents (e.g., quotes, ReTweet), and consider only raw textual data.

\section{Simulating novelty in textual data}\label{simulation}

In this section, we see how to generate textual data corresponding to our novelty typology constructed in the previous section. Then, we explain how we use a mixture model to simulate these data. These textual data are simulated using the concept of topic: namely, a probability distribution over the vocabulary.

\subsection{General framework}

\begin{figure}[h!]
    \centering
        \includegraphics[width=\textwidth]{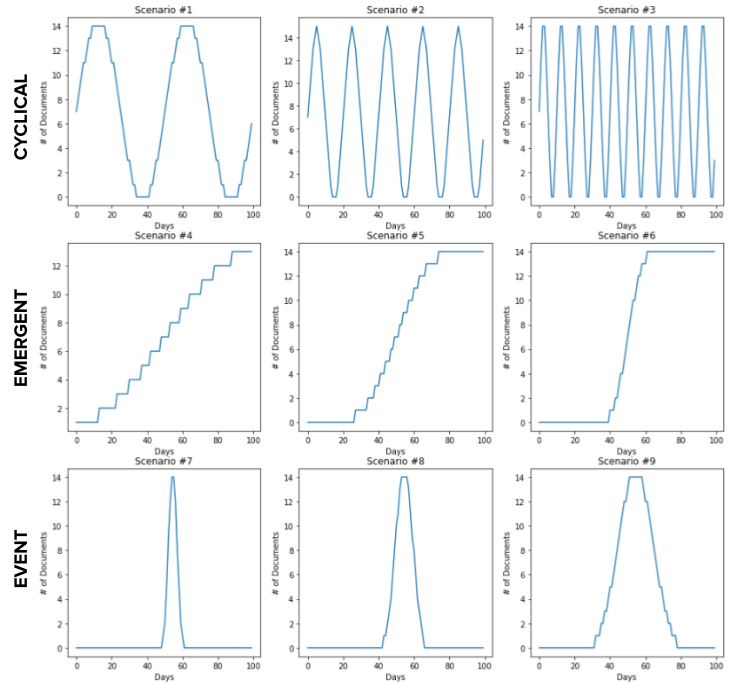}
        \caption{\label{fig:signal} 9 different scenarios of novelty dynamics over time.}
    \end{figure}

    Our main goal is to test the ability of different algorithms to detect novelty in textual data streams.
    We propose a precise methodology that allows us to measure the influence of various parameters on the results. In this part, we explain how we set up the datasets to create different novelty dynamics. In order to account for different novelty types, we build different scenarios. These scenarios correspond to three main types: cyclical, emergent and event. For each type, we vary some parameters to study the algorithms' performances. These scenarios are shown in Figure \ref{fig:signal}. To measure the sensitivity of the algorithm, we define a dataset as a set of topics: one novel topic has a temporal dynamics defined in one of the scenarios of Fig.~\ref{fig:signal} and the 9 others are constant over time (our ``normal'' data).
    
\subsection{Simulating}

To simulate textual data, we use a mixture model\footnote{The code for simulation is available at \url{https://github.com/clechristophe/NoveltySimulator}} \cite{nigam2000text}. For each document $d$ we assign a topic $z$ and then sample 100 words from its probability distribution over the 10,000 words of the vocabulary. It should be noted that we use a bag-of-words approach and that the order with which our words are generated is not important. We do not really simulate documents strictly speaking but rather sets of bag-of-words that are drawn from the topic distribution.
The mixture model that allows us to generate these probability distributions depends on a hyperparameter $\alpha$ that controls the sparsity of the vocabulary distribution. It gives us the ability to generate topics that are more or less close in terms of Kullback-Leibler Divergence (KL-Div). This is an important feature to test the different algorithms. When using topic models or probability distributions over a vocabulary, it is common to use the Kullback-Leibler Divergence to have an idea of the proximity between two topics.\\

We have simulated probability distributions on the vocabulary for each topic and assigned one topic per document. This allows us to create a simulated dataset.
To add the temporal dimension, we sort the documents in the desired order. This order is given by the dynamics in Figure \ref{fig:signal}. These scenarios give us the ability to test how an algorithm performs for each of them.

\section{Experiments}\label{experiments}
    
    In this section, we present the algorithms we study, the tasks they originally solve and how we adapt them for our evaluation framework. 

    \subsection{Tasks to resolve}

    In order to evaluate the methods originally designed for different applications, we define several tasks to be solved, with evaluation measures specific to each of these tasks.
    We measure the performance of the different methods on three tasks:
    \begin{itemize}
        \item Task 1: The goal is to raise an alert as soon as the novelty appears in the dataset. The evaluation measure corresponds to the delay between the day of novelty appearance  and the day of the alert.\\
        \item Task 2: The goal is to detect words associated with novelty. As the topics in our documents are entirely simulated, we know the words associated with the novel topic. By using the 100 most likely words of the novel topic, we build a ground truth and we can use precision/recall measures.\\
        \item Task 3: The goal is to detect the documents responsible for the appearance of novelty. Thanks to our simulation, we know the topic associated with each document. We have a ground truth and we can also use precision/recall measures for the evaluation.
    \end{itemize}
    
    \subsection{Evaluated Algorithms}

    The chosen algorithms come from different application fields: some are used for event detection, especially on Twitter, others for the detection of first stories, and others to observe the evolution of the language over time. In this section, we explain the main idea behind each of these methods, which task it is expected to solve in the first place, and finally how we adapt it to solve other tasks related to novelty detection.\\
    
    \textbf{Detections, bounds, and timelines: Umass and tdt-3 \cite{allan2000detections}} (\textbf{TF-IDF}): this paper is among the first in the field of First-Story-Detection. This method represents the documents in a space corresponding to the TF-IDF score associated to document words. Then it uses a k-nearest-neighbor-based search with the cosine dissimilarity (1-cos) to identify new documents. This work makes the assumption that a document appearing far from its nearest neighbors is new. \\
        
 \textbf{Structured event retrieval over microblog archives \cite{metzler2012structured}}(\textbf{BS}): this paper focuses on the detection of sudden bursts in Twitter data. It develops a metric named \textit {Burstiness Score} based on the word frequencies over each day. To detect new words, we consider the words with the most important BS score every day and compare them to our ground truth of words coming from our simulated topics.\\
        
\textbf{Towards effective event detection, tracking and summarization on microblog data \cite{long2011towards}}(\textbf{DF}): this paper uses the \textit{Document Frequency} feature and a Jaccard distance to form ``new words clusters'' every day that we compare with our ground truth.\\
        
\textbf{On-line trend analysis with topic models: twitter trends detection topic model online \cite{lau2012line}}(\textbf{OLDA}): this paper presents a method that detects events over time slices. It is a variant of LDA that updates the proportion of word/topic every day. The new topics are identified via a Jensen-Shannon distance with the topics of the previous days. For word detection, we can compare the most likely words of a topic identified as new with our ground truth.\\
        
\textbf{Topicsketch: Real-time bursty topic detection from twitter \cite{xie2016topicsketch}} (\textbf{TopicSketch}): this paper presents a method adapted to event detection on Twitter data. The proposed model monitors the speed and acceleration of word frequencies to detect novelty. Every day, the model can raise an alert for one or more words that we can compare to our ground truth.

    \input{table/algo.tex}
    
    Table \ref{tab:tasks} gives us information about what tasks the algorithms are originally designed to solve. In order to evaluate these methods, we need to adapt them for other tasks and measure their performances on these new tasks. Even if the adaptation is not optimal and does not represent the best way to resolve a specific task, it gives us insight on how the method or feature or even just the representation helps. Now, we describe how we adapt these algorithms. \\
    
    In order to adapt the TF-IDF \cite{allan2000detections} method to new word detection, we use the IDF of each word (traditionally associated with rarity) to see if the most important values of each day is a good indicator of novelty. For the method based on the Burstiness Score \cite{metzler2012structured}, we aggregate the burstiness score of each word in the document to create a burstiness score for the document. This aggregation can be done in several way (mean, median, percentile). To use the Document Frequency \cite{long2011towards} feature to detect new documents, we were inspired by \cite{allan2000detections} and we use a search for k-nearest neighbors in a representation space for documents built from their \textit{Document Frequency}. When using OLDA \cite{lau2012line} to detect new topics, we compare the documents belonging the most to the topic identified as novel with our ground-truth documents. The TopicSketch \cite{xie2016topicsketch} method is used to raise alerts when a term frequency is detected as abnormal. For the document detection, we observe the documents that are responsible for the alert as well as previous documents containing the detected term.

    \section{Results}\label{results}

    In this section, we present the performance results of the evaluated algorithms on the tasks defined earlier. First, we enumerate specific results that help us determine the ideal parameters and understand their influence on the performance. We present these results on specific scenarios and trace the temporal evolution of precision/recall measures over time as well as the evolution of the number of new documents. Finally, we test each of these methods on an excerpt of the New York Times Annotated Dataset.

    \subsection{Influence of the distance between topics}
    
\begin{figure}
    \centering
    \subfigure
    {
        \includegraphics[width=.45\textwidth]{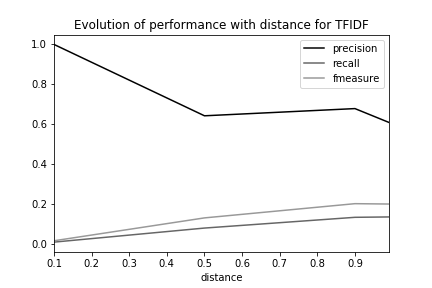}
        \label{fig:tfidf_dist}
    }
    \\
    \subfigure
    {
        \includegraphics[width=.45\textwidth]{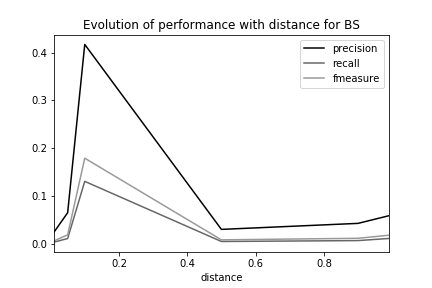}
        \label{fig:bs_dist}
    }
    \subfigure
    {
        \includegraphics[width=.45\textwidth]{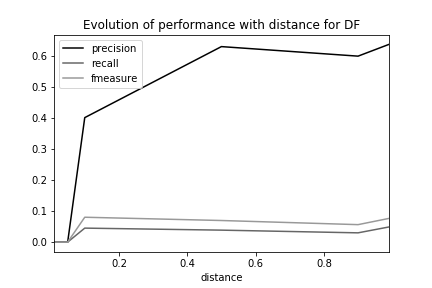}
        \label{fig:df_dist}
    }
        \subfigure
    {
        \includegraphics[width=.45\textwidth]{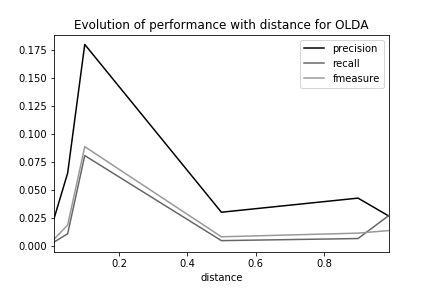}
        \label{fig:olda_dist}
    }
        \subfigure
    {
        \includegraphics[width=.45\textwidth]{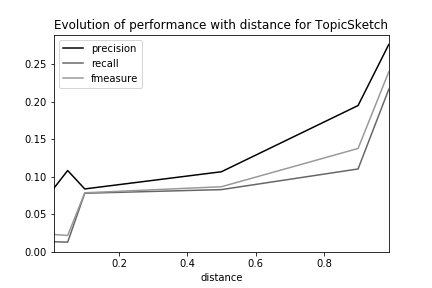}
        \label{fig:sketch_dist}
    }
    \caption{Evolution of performance compared to the divergence between topics (as calculated with KL-div) for each tested algorithm.}
    \label{fig:dist}
\end{figure}

    First we study the influence of the distance between topics on the performance of novelty detection algorithms. As a reminder, we generated datasets where we control the Kullback-Leibler divergence (KL-Div) between topics. We generated 6 datasets with the following KL-Div values: $0.01,0.05,0.1,0.5,09,0.99$. Our intuition is the following: since most algorithms are based on similarity measure between documents and/or on the evolution of term frequencies in the corpus, we assume that novelty detection becomes more difficult with the reduction of the distance between topics. We test this hypothesis by observing the evolution of the performance on the document detection task. We see on Figure~\ref{fig:dist} that our hypothesis is globally true, especially for the two research-based methods compared to nearest neighbors and TopicSketch. Finally, it is interesting to note that, although the detection level is very low for the DF and OLDA methods, they both have optimal results for KL-Div $= 0.1$.
    
    \subsection{Influence of the choice of k for knn-based algorithms}
    
\begin{figure}
        \centering
        \subfigure
        {
            \includegraphics[width=.45\textwidth]{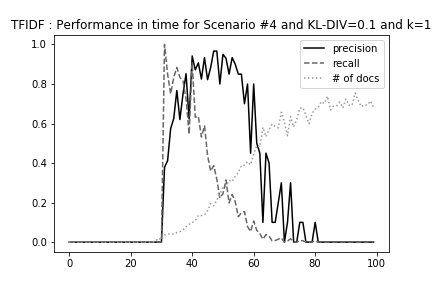}
            \label{fig:tfk1}
        }
        \subfigure
        {
            \includegraphics[width=.45\textwidth]{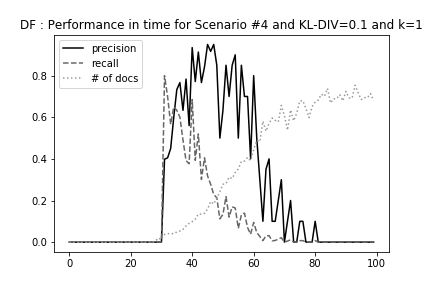}
            \label{fig:dfk1}
        }
        \subfigure
        {
            \includegraphics[width=.45\textwidth]{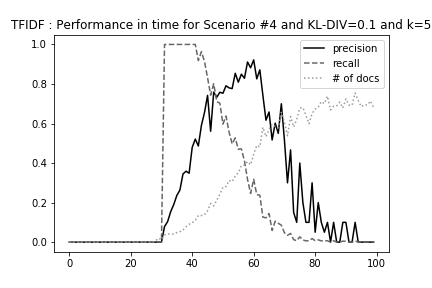}
            \label{fig:tfk5}
        }
        \subfigure
        {
            \includegraphics[width=.45\textwidth]{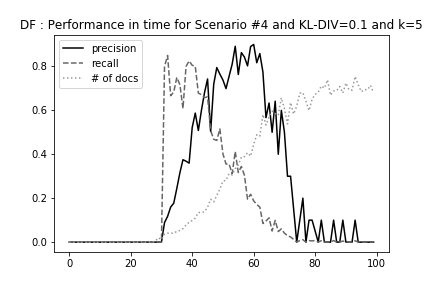}
            \label{fig:dfk5}
        }
        \subfigure
        {
            \includegraphics[width=.45\textwidth]{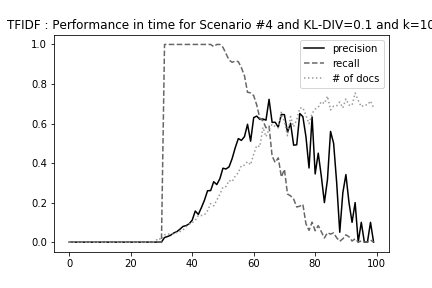}
            \label{fig:tfk10}
        }
        \subfigure
        {
            \includegraphics[width=.45\textwidth]{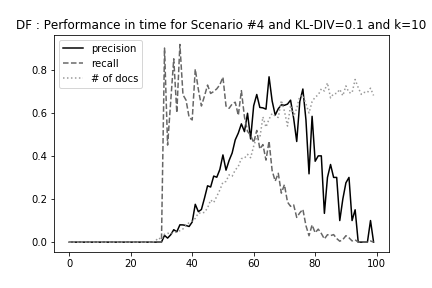}
            \label{fig:dfk10}
        }
        \caption{Evolution of Precision and Recall measure with the appearance of novelty depending on the value of k. 
        The curve representing the number of documents per day is normalized.}
        \label{fig:kanalysis}
    \end{figure}

    Several tested algorithms are based on a k-nearest-neighbors search. The choice of the value of the best k can greatly influence the results.
    Actually, the method of Allan \cite{allan2000detections} is originally thought to carry out a first story detection task, i.e. detect the first document far from all its neighbors in the representation space.
    The aggregation of several similar documents in the same subspace can be interesting to detect.
    It is necessary to modify the value of k to observe this type of phenomenon. Since we aim at correctly classifying the maximum number of novel documents, we focus on recall.
    We observe on Figure~\ref{fig:kanalysis} that, the higher the value of k, the longer good performances last over time, which is what we expect.
    Actually, the new documents appear in the same part of the representation space and the performance starts to decrease as soon as $k$ documents have appeared. This is what we can expect for a first story or anomaly detection method.
    
    \subsection{Influence of the speed of novelty appearance}

    Some methods that we are evaluating in this work have been initially developed to address event detection. Events are associated to topics that appear very quickly and in large quantities before disappearing almost as quickly.
    It is a typical behavior we can observe in Twitter data. This idea corresponds to the scenarios 7,8 and 9 that we simulate in our data. Here we evaluate the influence of the speed at which new topics appear on two methods (OLDA and TopicSketch) on the document detection task.
    
    \begin{figure}[h!]
    \centering
        \includegraphics[width=0.8\textwidth]{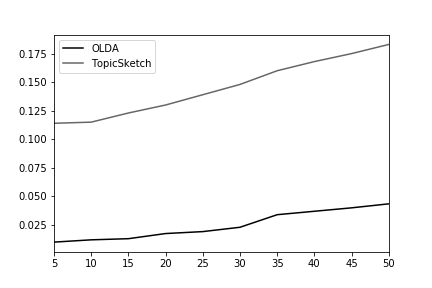}
        \caption{\label{fig:slope} Evolution of F-Measure with the slope parameter of the novelty appearance.}
    \end{figure}
    
    We observe on Figure \ref{fig:slope}, as expected, that the two tested methods are more sensitive to a steep slope and therefore better detect the novelty when it appears quickly. We can conclude that these methods are more suited to event detection tasks than to the detection of novelty appearing slowly over time.
        
    \subsection{General results}
    
    In this section, we present the general results for the detection of new words (Task 2) and new documents (Task 3).

\input{table/task_2.tex}

    On Figure~\ref{tab:genresults}, we see that, despite the overall low results, it is still the method based on TopicSketch that achieves the most competitive performance. For all the measures relating to Task 2, the latter is better. We also see that the OLDA method does not work at all for detecting new words in the case of cyclical scenarios. Finally, it is generally the recall that is very weak, which means that few words really new are correctly found by the algorithms. In terms of document detection, the results are more diverse, only the TopicSketch and TF-IDF methods stand out. We note that the two research-based methods compared to the nearest neighbors (TF-IDF \& DF) work very poorly on the cyclical scenarios, which is in line with our initial intuition. This table shows that the results are quite low but it also highlights the fact that tested algorithms work differently on different scenarios.
    
    \subsection{Evaluation on the New York Times Annotated Dataset}

\begin{figure}[h!]
    \centering
        \includegraphics[width=\textwidth]{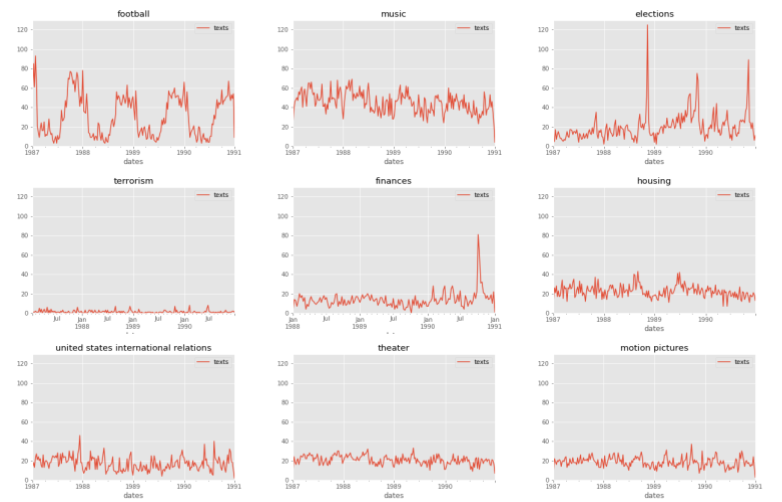}
        \caption{\label{fig:nyt} Temporal evolution of some categories in the NYT.}
    \end{figure}

    To illustrate the performances of the different algorithms, we tested them on a real dataset\footnote{\url{https://catalog.ldc.upenn.edu/LDC2008T19}} corresponding to articles published by the New York Times. It is a dataset of 1.8 million items (20 years of data) manually annotated by topic. We decided to carry out our experiments on the first 4 years and on topics with specific temporal behaviors.
    In order to evaluate our methods, we need to artificially incorporate novelty into our data. For each of the topics, whose temporal evolution is presented in Figure~\ref{fig:nyt}, we completely deleted its associated documents in 1987. This year of data thus forms our historic. For the next three years (1988-1989-1990), the topic is present in the data and the documents associated with it forms our ground truth.

    We now test the ability of different algorithms to detect the targeted ``new'' topic.
    The representation spaces (TF-IDF, DF) or topics (OLDA) are initialized with data containing no text of the targeted topic. We then launch the methods \emph{with} the documents corresponding to the new topic. This way, we have a ground truth and we know what we want to detect. For evaluation measures, that we present below, we use Area Under Curve (AUC) to check whether the algorithms have a good tendency to classify the really novel documents first. We use AUC instead of precision-recall because we find it more relevant for this real dataset. Contrary to the simulated data, we have not such a strong ground truth and AUC provides a more relevant summary of the method ability to rank (tagged) novel documents before other (less relevant) documents. An AUC score close to 1 is better.
        
    \begin{table}[]
    \centering
    \begin{tabular}{|l|c|c|c|c|c|}
    \hline
                & Football & Finances & Theater & Art   & Elections \\ \hline
    TF-IDF      & 0,257    & 0,292    & \textbf{0,744}   & 0,514 & 0,334     \\ \hline
    BS          & 0,281    & 0,312    & 0,412   & 0,474 & 0,51      \\ \hline
    DF          & 0,241    & 0,242    & 0,614   & \textbf{0,534} & 0,378     \\ \hline
    OLDA        & 0,294    & 0,411    & 0,531   & 0,521 & 0,671     \\ \hline
    TopicSketch & \textbf{0,641}    & \textbf{0,524}    & 0,514   & 0,531 & \textbf{0,751}     \\ \hline
    \end{tabular}
    \caption{AUC scores on some categories of the NYT Annotated Corpus.}
    \end{table}

    \begin{table}[]
    \centering
    \begin{tabular}{|l|c|c|c|c|c|}
    \hline
                & USA intl. relations & Music & Motion Pictures & Housing & Terrorism \\ \hline
    TF-IDF      & 0,301              & \textbf{0,588} & 0,675           & 0,300     & 0,553     \\ \hline
    BS          & 0,14               & 0,201 & 0,254           & 0,145   & 0,341     \\ \hline
    DF          & 0,286              & 0,514 & 0,612           & 0,241   & 0,513     \\ \hline
    OLDA        & 0,561              & 0,54  & \textbf{0,712}           & \textbf{0,610}    & 0,491     \\ \hline
    TopicSketch & \textbf{0,614}             & 0,485 & 0,631           & 0,524   & \textbf{0,540}      \\ \hline
    \end{tabular}
    \caption{AUC scores on some categories of the NYT Annotated Corpus.}
    \end{table}

    Note that each method does not necessarily perform well on the same topics.
    For example, only the TopicSketch method achieves interesting results for \textit{Football}. This is a typical cyclical topic and therefore the words associated with these events occur a lot in the corpus, especially during matches such as the Superbowl. Some topics like \textit{Terrorism} are more complicated to detect. This is explained by the fact that they still remain at the weak signal level in the corpus.
    
\section{Conclusion and future works}\label{conclusion}

In this work, we have built a common framework for the task of novelty detection. We have shown that several works of the literature studied this field but not necessarily from the same angle: the novelty was not defined the same way, the tasks were not identical and no common benchmark exists in the literature. We proposed our own definition of novelty with different novelty types. Working on simulated data allowed us to control the data creation process and, thus, to better measure the performances of the different methods according to each situation. Although artificial data was useful for challenging the chosen algorithms when confronted to archetypal situations, attention should be paid on the conclusion we can draw from the experiments.
Actually some concerns may be raised about the global low scores we achieved.
This can be due to a simplified data generation
process, the adaption of the existing methods designed for solving close but different tasks, or the choice we made for the competing algorithms.

This work allowed us to draw conclusions on the features and metrics to use according to the novelty type we aim to detect. Actually, methods using metrics based on a simple count of words, like TopicSketch (speed and acceleration) \cite{xie2016topicsketch}, seem to be able to detect novelty more easily than methods based on a representation space of words or documents
Even if the results are globally low, the goal of this paper was to show that: a) there is no clear consensus around the definition of novelty, b) a method designed for one scenario (e.g., event detection) may be adapted to solve another scenario (e.g., emergent trend identification), c) different novelty scenarios need different types of methods, which is related to the well-known No Free Lunch theorem.

However, working with simulated bag-of-word data prevents us from using context information and therefore using embeddings-based methods. In the future, we will focus on these methods applied to novelty detection and we also plan to evaluate methods based on more recent topic modeling models. Also, some more complex novelty types are interesting to study such as the disappearance of a topic or the fusion of two topics.

\bibliographystyle{abbrv}
\bibliography{samplepaper}

\newpage
\appendix 
\end{document}

%% file: table/algo.tex
\begin{table} \centering
\begin{tabular}{|r|ccc|}
Algorithm & Task 1 & Task 2 & Task 3\\
\midrule
TF-IDF  &  &  & X\\
BS  &  & X & \\
DF  &  & X & \\
OLDA &  & X & \\
TopicSketch & X & X & \\ \bottomrule
\end{tabular}
\caption{Which tasks the algorithms are originally designed to solve?} \label{tab:tasks} 
\end{table}

%% file: table/task_2.tex
\begin{table}
\centering
\begin{tabular}{|l|l|l|l|l|l|l|l|l|l|} 
\hline
\multicolumn{10}{|c|}{\textbf{Results for task 2 : Detection of new words}} \\
\hline
\hline
\multirow{2}{*}{Methods} & \multicolumn{3}{l|}{Scenario 1}                           & \multicolumn{3}{l|}{Scenario 2}                           & \multicolumn{3}{l|}{Scenario 3}                            \\ 
\cline{2-10}
                         & P                 & R                 & F                 & P                 & R                 & F                 & P                 & R                 & F                  \\ 
\hline
TF-IDF                   & 0.073          & 0.005          & 0.010          & 0.085          & 0.004          & 0.009         & 0.089          & 0.006          & 0.012           \\ 
\hline
BS        & 0.044          & 0.004          & 0.007          & 0.025          & 0.006          & 0.010          & 0.007          & 0.009          & 0.008           \\ 
\hline
DF      & 0.186          & 0.005          & 0.010          & 0.179          & 0.004          & 0.008          & 0.205          & 0.006          & 0.013           \\ 
\hline
OLDA                     & 0                 & 0                 & 0                 & 0                 & 0                 & 0                 & 0                 & 0                 & 0                  \\ 
\hline
TopicSketch              & \textbf{0.239} & \textbf{0.106} & \textbf{0.138} & \textbf{0.286} & \textbf{0.104} & \textbf{0.153} & \textbf{0.281} & \textbf{0.106} & \textbf{0.151}  \\
\hline
\hline
\multirow{2}{*}{Methods} & \multicolumn{3}{l|}{Scenario 4}                            & \multicolumn{3}{l|}{Scenario 5}                            & \multicolumn{3}{l|}{Scenario 6}                             \\
\cline{2-10}
                         & P                 & R                 & F                 & P                 & R                 & F                 & P                 & R                 & F                  \\\hline
TF-IDF                   & 0.247          & 0.006          & 0.012          & 0.400          & 0.003          & 0.007          & 0.496          & 0.009          & 0.017           \\\hline
BS                       & 0.173          & 0.005          & 0.010          & 0.308          & 0.003          & 0.006          & 0.375          & 0.010          & 0.020           \\\hline
DF                       & 0.330          & 0.005          & 0.011          & 0.504          & 0.003          & 0.006          & 0.594          & 0.008          & 0.017           \\\hline
OLDA                     & 0.130          & 0.006          & 0.011          & 0.197          & 0.002          & 0.005          & 0.420          & 0.010          & 0.021           \\\hline
TopicSketch              & \textbf{0.483} & \textbf{0.105} & \textbf{0.172} & \textbf{0.552} & \textbf{0.103} & \textbf{0.170} & \textbf{0.716} & \textbf{0.109} & \textbf{0.187}  \\
\hline
\hline
\multirow{2}{*}{Methods} & \multicolumn{3}{l|}{Scenario 7}                           & \multicolumn{3}{l|}{Scenario 8}                           & \multicolumn{3}{l|}{Scenario 9}                            \\ 
\cline{2-10}
                         & P                 & R                 & F                 & P                 & R                 & F                 & P                 & R                 & F                  \\ 
\hline
TF-IDF                   & 0.456          & 0.008          & 0.015          & 0.466          & 0.005          & 0.011          & 0.499          & 0.010          & 0.021           \\ 
\hline
BS                       & 0.352          & 0.008          & 0.017          & 0.366          & 0.005          & 0.010          & 0.377          & 0.011          & 0.021           \\ 
\hline
DF                       & 0.543          & 0.007          & 0.015          & 0.604          & 0.006          & 0.012          & 0.621          & 0.010          & 0.020           \\ 
\hline
OLDA                     & 0.302          & 0.009          & 0.017          & 0.248          & 0.006          & 0.013          & 0.341          & 0.010          & 0.022           \\ 
\hline
TopicSketch              & \textbf{0.608} & \textbf{0.108} & \textbf{0.182} & \textbf{0.641} & \textbf{0.106} & \textbf{0.179} & \textbf{0.762} & \textbf{0.111} & \textbf{0.193}  \\
\hline

\hline

\multicolumn{10}{|c|}{\textbf{Results for task 3 : Detection of new documents}} \\
\hline
\hline
\multirow{2}{*}{Methods} & \multicolumn{3}{l|}{Scenario 1}                           & \multicolumn{3}{l|}{Scenario 2}                           & \multicolumn{3}{l|}{Scenario 3}                            \\ 
\cline{2-10}
                         & P                 & R                 & F                 & P                 & R                 & F                 & P                 & R                 & F                  \\ 
\hline
TF-IDF                   & 0                 & 0                 & 0                 & 0.062          & 0.001          & 0.003          & 0.093          & 0.002          & 0.005           \\ 
\hline
BS                       & 0.028          & 0.005          & 0.008          & 0.083          & 0.013          & 0.023          & \textbf{0.125} & 0.018          & 0.032           \\ 
\hline
DF                       & 0                 & 0                 & 0                 & \textbf{0.193} & 0.052          & \textbf{0.071} & 0                 & 0                 & 0                  \\ 
\hline
OLDA                     & 0.029          & 0.005          & 0.008          & 0.084          & 0.014          & 0.024          & 0.109          & 0.018          & 0.029           \\ 
\hline
TopicSketch              & \textbf{0.051} & \textbf{0.036} & \textbf{0.040} & 0.062          & \textbf{0.056} & 0.043          & 0.096          & \textbf{0.078} & \textbf{0.083}  \\
\hline
\hline
\multirow{2}{*}{Methods} & \multicolumn{3}{l|}{Scenario 4}                           & \multicolumn{3}{l|}{Scenario 5}                           & \multicolumn{3}{l|}{Scenario 6}                            \\ 
\cline{2-10}
                         & P                 & R                 & F                 & P                 & R                 & F                 & P                 & R                 & F                  \\ 
\hline
TF-IDF                   & 0.600          & 0.023          & 0.04           & \textbf{0.825} & 0.055          & 0.103          & \textbf{0.856} & 0.046          & 0.085           \\ 
\hline
BS                       & 0.048          & 0.005          & 0.010          & 0.102          & 0.011          & 0.019          & 0.258          & 0.027          & 0.049           \\ 
\hline
DF                       & \textbf{0.638} & \textbf{0.026} & 0.050          & 0.650          & 0.029          & 0.056          & 0.741          & 0.046         & 0.087           \\ 
\hline
OLDA                     & 0.048          & 0.005          & 0.010          & 0.077          & 0.011          & 0.019         & 0.108          & 0.027          & 0.043           \\ 
\hline
TopicSketch              & 0.150          & 0.096          & \textbf{0.114} & 0.165          & \textbf{0.123} & \textbf{0.139} & 0.229          & \textbf{0.155} & \textbf{0.183}  \\
\hline
\hline
\multirow{2}{*}{Methods} & \multicolumn{3}{l|}{Scenario 7}                           & \multicolumn{3}{l|}{Scenario 8}                  & \multicolumn{3}{l|}{Scenario 9}                            \\ 
\cline{2-10}
                         & P                 & R                 & F                 & P                 & R        & F                 & P                 & R                 & F                  \\ 
\hline
TF-IDF                   & \textbf{0.818} & 0.278          & 0.375          & \textbf{0.843} & 0.167 & \textbf{0.259} & \textbf{0.852} & 0.085          & 0.149           \\ 
\hline
BS                       & 0.604          & \textbf{0.481} & \textbf{0.536} & 0.326          & 0.133 & 0.189          & 0.170          & 0.034          & 0.057           \\ 
\hline
DF                       & 0.770          & 0.054          & 0.100          & 0.710         & 0.071 & 0.122          & 0.838          & 0.052          & 0.090           \\ 
\hline
OLDA                     & 0.104          & 0.181          & 0.132          & 0.069          & 0.080 & 0.070          & 0.061          & 0.045          & 0.038           \\ 
\hline
TopicSketch              & 0.216          & 0.186          & 0.197          & 0.209          & 0.128 & 0.150         & 0.221          & \textbf{0.142} & \textbf{0.166}  \\
\hline
\end{tabular}
\caption{Precision (P), Recall (R) and F-Measure (F) results for each evaluated algorithms on 9 scenarios for Tasks 2 \& 3.}
\label{tab:genresults}
\end{table}